# WinNet: Make Only One Convolutional Layer Effective for Time Series Forecasting

Wenjie Ou[1], Zhishuo Zhao[1], Dongyue Guo[1,2], Zheng Zhang[1], Yi Lin[1,2(✉)]

[1] School of Computer Science, Sichuan University, Chengdu, China
[2] National Key Laboratory of Fundamental Science on Synthetic Vision, Sichuan University, Chengdu, China
`yilin@scu.edu.cn`

**Abstract.** Deep learning models have recently achieved significant performance improvements in time series forecasting. We present a highly accurate and simply structured CNN-based model with only one convolutional layer, called WinNet, including (i) Sub-window Division block to transform the series into 2D tensor, (ii) Dual-Forecasting mechanism to capture the short- and long-term variations, (iii) Two-dimensional Hybrid Decomposition (TDD) block to decompose the 2D tensor into the trend and seasonal terms to eliminate the non-stationarity, and (iv) Decomposition Correlation Block (DCB) to leverage the correlation between the trend and seasonal terms by the convolution layer. Results on eight benchmark datasets demonstrate that WinNet can achieve SOTA performance and lower computational complexity over CNN-, MLP- and Transformer-based methods.

## 1. Introduction

Time series forecasting (TSF) has been widely used in the prediction of energy consumption, transportation, economic planning, weather and disease transmission. Deep learning models have recently achieved significant performance in prediction. Benefiting from the self-attention, Transformer-based methods [1, 2, 3, 4, 8, 11] can capture the long-term temporal dependency and achieve SOTA performance. However, they are not sensitive to the periodicity of time series and have high computational complexity. Therefore, many MLP-based methods [5, 9, 10, 13] have been proposed to achieve better predictions while balancing efficiency.

Periodicity in time series data is the inherent property for discovering temporal dependencies in the TSF tasks. TimesNet [6] converts the sequence into a two-dimensional (2D) tensor by multiple periods. Secondly, time series decomposition has been proposed in Autoformer [3], yet the correlation between the terms obtained by decomposition has not received much attention. Based on the lagged-correlation analysis, we can conclude that there is an extremely strong lag correlation between the trend and seasonal terms, and the correlation has a periodic pattern, as shown in **Fig. 1**. They combine the terms from the time series decomposition strategy by simply adding instead. Finally, TimesNet [6] calculates the top-k periods of the sequence obtained from the Fast Fourier Transform, divides the same sequence into multiple 2D tensors and models them separately to extract the features of the changes in each period. This increases the time complexity and memory overhead, which inspires us to reconsider the CNN approaches for the TSF tasks.

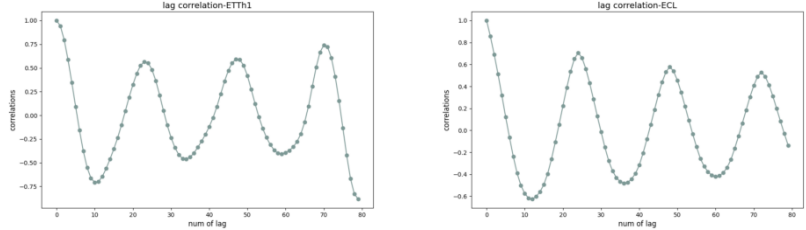

**Fig. 1.** The lag correlation between the trend and seasonal terms in ETTh1 and ECL datasets. We can see that the lag correlation is very strong and that there is a periodic pattern of 24.

Motived by the above observations, we propose a simple and effective **Win**dow-enhanced CNN-based model with only one convolutional layer for the TSF tasks, called **Win**Net. To eliminate the non-stationarity of the time series, we propose a two-dimensional hybrid decomposition (TDD) block for the 2D tensor. The TDD block is proposed to decompose the 2D tensor into trend and seasonal terms, which balances the short- and long-term variations. To mine the correlation between the two terms, the DCB block is innovatively designed to combine the two terms by the convolutional layer. From the result, WinNet can reduce the Mean Squared Error and Mean Absolute Error by 20.3% and 13.4%, compared to TimesNet.

In summary, this work contributes to the TSF tasks in the following ways:

- Only one vanilla CNN layer is designed as the backbone of the prediction network, which notably reduces the training memory and computational complexity and improves experimental efficiency. This indicates that the vanilla CNN model can be more effective than most well-designed models for the TSF tasks.
- To eliminate the non-stationarity, we propose a two-dimensional hybrid decomposition block. The block extracts local variations within- and cross-window by a sliding window with the moving average method. We apply the TDD block to various deep models and boost the performance by a large margin.
- Extensive experiments are conducted on 8 benchmark datasets across multiple domains. Our experimental results demonstrate that WinNet outperforms other comparative baselines in both univariate and multivariate prediction tasks. WinNet can achieve better prediction accuracy and efficiency for the TSF tasks.

## 2.    Related Work

### 2.1    CNN Network for Time Series Forecasting

CNN are also used in Computer Vision, where they can be used to extract local features from images. Recently, CNN-based methods [6, 7, 12] have been proposed to mine the periodicity of time series. SCINet [12] obtains features at different time resolutions using a multi-layer binary tree structure. In TimesNet [6], the Inception network is applied to process the temporal points.

Furthermore, a multi-scale isometric convolutional network with multi-scale branches is proposed in MICN [7] to capture both local and global features of the temporal sequence.

### 2.2 Time Series Decomposition for Time Series Forecasting

Time series data always tends to be non-stationary, and relies largely on the model's predictive ability. To solve the problem, time series decomposition is applied to decompose the sequence into the trend and seasonal terms to enhance the non-stationarity in the time series. Autoformer [3] and FEDformer [2] eliminate the trend terms and focus on seasonal terms modeling. Furthermore, DLinear [5] and PatchTST [4] models the trend and seasonal terms separately by two same backbones to obtain the predictor terms. However, we find out that the correlation between the two terms obtained by decomposition have not received much attention.

## 3. Framework of WinNet

The framework of WinNet is shown in **Fig. 2**, which aims to learn a map between the input $X_{in} \in R^{sl \times c}$ and output $X_{out} \in R^{pl \times c}$, where sl, pl, c are the lengths of input, prediction and feature.

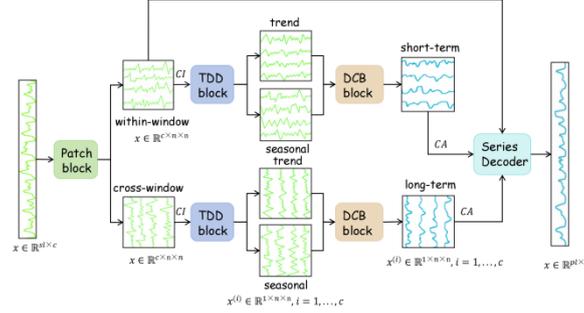

**Fig. 2.** The model architecture of WinNet.

### 3.1 Sub-window Division

For the given input data $X_{in} \in R^{sl \times c}$, we perform a linear mapping layer along the temporal dimension. Moreover, we transform the input length into a 2D tensor. Subsequently, the sequence is divided into n sub-windows $X_{in} \in R^{n \times n \times c}$. After division, we get a 2D tensor, where the temporal points within rows are in chronological order (short-term variation), and its columns are arranged across a fixed time interval (long-term variation). We employ RevIN [15] to solve the distribution shift in time series data. Unfold(·) operation is the patching process by the Reshape(·) function. The process is shown as:

$$X_{in} = Linear(Permute(RevIN(X_{in})))$$
$$X_w = Unfold(X_{in}, size = n, stride = n) \qquad (1)$$

### 3.2     Dual-Forecasting Heads

BiLSTM is early to adopt the two-head mechanism, which can better capture bidirectional information dependencies. Motivated by the mechanism, we propose a dual-forecasting heads for WinNet to capture the features of the short- and long-term variations, respectively.

$$X_{cw} = Transpose(X_w) \tag{2}$$

By transposing the tensor within-window $X_w$, we get a 2D tensor cross-window $X_{cw}$, whose row highlights the long-term variation. We feed the two tensors into the model in parallel and capture features of long and short variations from different perspectives.

### 3.3     Two-dimensional Hybrid Decomposition

In general, existing methods for time series decomposition mainly focused on decomposing the 1D sequence. Inspired by the series decomposition idea in DLinear [5], we propose a Two-dimensional Hybrid Decomposition block. Specifically, a trend-padding operation is dedicatedly designed to perform the convolutional operation at the boundary. As shown in **Fig. 3**, after the operation, the 2D tensor $X \in R^{n \times n \times c}$ are padded into a new 2D tensor $X \in R^{(n+p) \times (n+p) \times c}$ where p means the padding size in row and column.

$$X_i = Trendpadding(X_i), i \in (w, cw)$$
$$X_i^t = Avgpool2D(X_i), i \in (w, cw) \tag{3}$$
$$X_i^s = X_i - X_i^t, i \in (w, cw)$$

where $X_i^t$ and $X_i^s$ are the trend and seasonal terms by decomposition. The Avgpool2D($\cdot$) is applied to extract both the features of the within-window (short-term variation) and cross-window (long-term variation). We apply the 2D decomposition idea to various mainstream deep models and boost the performance by a large margin.

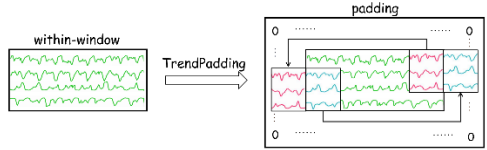

**Fig. 3.** The figure of Trendpadding. Unlike the 0 or the same padding mode in common CNNs, the neighbor samples (before or after) in the original sequence are selected as the padding item.

### 3.4     Decomposition Correlation Block

Previous models [5, 4, 7] fail to capture the correlation between the trend and seasonal terms. Based on the lagged-correlation analysis, we find a lag correlation between the trend and seasonal terms, and the two terms can both influence the variations of future temporal points. Moreover, the current temporal point of the seasonal term is strongly correlated with future temporal points of the trend term at fixed intervals. The correlation between the two terms has implications for future temporal points. As shown in **Fig. 4**, in the within-window $X_w$, each row extracted by the convolution kernel represents short-term variation and the column represents long-term variation. The short- and long-term variation of the two terms can be simultaneously extracted by one convolutional kernel, and the learned parameters can perform a proportional aggregation of them to learn the correlation between the two terms $X_w^t, X_w^s$. We choose CNN as our backbone network to synthesize the local correlation of the current temporal point in the window neighborhood. The process is described below:

$$\bar{X}_i = Concat(CI(X_i^t, X_i^s)), i \in (w, cw)$$

$$\bar{X}_i = Sigmoid\left(ReLU(Conv2D(\bar{X}_i))\right), i \in (w, cw) \quad (4)$$

$$\bar{X}_i = CA(Dropout(\bar{X}_i)), i \in (w, cw)$$

where CI(·) and CA(·) mean channel independence and aggregation strategy. In the CI(·), we split both trend and seasonal terms at the channel dimension and concatenate them into two-channel tensors. We feed the tensors at the same channel into the Conv2D(·) function. The ReLU(·) and Sigmoid(·) are the activation functions for enhanced model learning. In the CA(·), we concatenate the single-channel output $\bar{X}_i$ after the DCB block at the channel dimension.

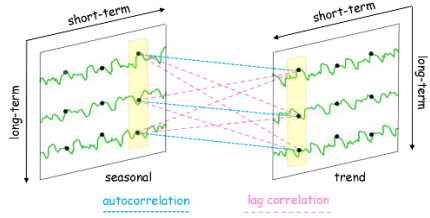

**Fig. 4.** There is an auto-correlation at corresponding positions and a lag correlation at the interleaved positions between the seasonal and trend terms.

### 3.5 Series Decoder

In this section, the Series Decoder block is designed to aggregate the local trend of the two tensors $X_{in}^w$ and $X_{in}^c$ for extracting global trend. Specifically, $\bar{X}_w$ is obtained by the within-window head, reflecting the short-term variation, and $\bar{X}_{cw}$ is obtained by the cross-window head, whose neighboring points in each row are spanned by the window size and reflect the long-term variation. This design can interactively learn the correlation of the short-period time steps with the long-term features, thus further extracting the global variation. The linear layer is designed to map the learned features into the prediction results. The whole process is:

$$X_{out} = \bar{X}_w + \bar{X}_{cw}$$

$$X_{out} = X_{out} + X_{in} \quad (5)$$

$$X_{out} = Permute(Linear(Reshape(X_{out})))$$

Table 1. The statistical feature details of eight datasets.

| Dataset | Size | Frequency | Length | Prediction Length |
|---|---|---|---|---|
| ETTm1&m2 | 7 | 15min | 69680 | {96, 192, 336, 720} |
| ETTh1&h2 | 7 | 1hour | 17420 | {96, 192, 336, 720} |
| Weather | 21 | 10min | 52696 | {96, 192, 336, 720} |
| ECL | 321 | 1hour | 26304 | {96, 192, 336, 720} |
| Traffic | 862 | 1hour | 17544 | {96, 192, 336, 720} |
| Exchange | 8 | 1day | 7588 | {96, 192, 336, 720} |

## 4.  Experiment

In this section, we conduct sufficient experiments on eight widely used datasets compared to state-of-the-art models to verify the effectiveness of WinNet.

### 4.1  Experimental Setup

**Datasets** A total of 8 real-world datasets are applied to validate the proposed approach. The details about these datasets are described in the **Table 1**.

**Baselines** To verify the effectiveness of WinNet for the TSF tasks, we choose several state-of-the-art models as the comparative baselines, including iTransformer [11], PatchTST [4], TimesNet [6], RLinear [10], DLinear [5].

**Implementation Details** All the models follow the same experimental setup with a prediction length of T = {96, 192, 336, 720} for the datasets. We set the input length L = 512. The Mean Squared Error (MSE) and Mean Absolute Error (MAE) are selected to measure the model performance for the TSF tasks. A smaller value indicates higher performance. All the experiments are implemented in PyTorch and conducted on a NVIDIA RTX4090 24GB GPU.

### 4.2  Main Results

In general, WinNet outperforms the baselines on multivariate and univariate forecasting tasks.
**Multivariate Results** As shown in **Table 2**, in general, WinNet essentially achieves the best performance for the listed datasets. Quantitatively, WinNet improves 22.5% in MSE and 14.5% in MAE compared to TimesNet [6], indicating that WinNet can more stably capture the long- and short-term variations. In addition, WinNet improves 10.1% in MSE and 6.8% in MAE against iTransformer [11], 8.3% in MSE and 5.2% in MAE against RLinear [10]. It is noted that WinNet achieves inferior performance on the traffic datasets due to the inability of the channel independence strategy to capture its multi-channel correlation.
**Univariate Results**  As shown in **Table 3**, WinNet significantly outperforms other SOTA models on almost all datasets, achieving an improvement of 14.3% in MSE and 11.5% in MAE for iTransformer [11], 13.3% in MSE and 8.5% in MAE for TimesNet [6], 20.8% in MSE and 14.9% in MAE for RLinear [10]. This demonstrates that the modules in WinNet indeed capture more useful information for univariate TSF tasks.

Table 2. Multivariate time series forecasting results. The input length is set as 512.

| Method | | WinNet | | TimesNet | | iTransformer | | PatchTST | | RLinear | | DLinear | |
|---|---|---|---|---|---|---|---|---|---|---|---|---|---|
| Metric | | MSE | MAE | MSE | MAE | MSE | MAE | MSE | MAE | MSE | MAE | MSE | MAE |
| ETTm1 | 96 | **0.283** | **0.335** | 0.377 | 0.397 | 0.309 | 0.363 | 0.293 | 0.346 | 0.309 | 0.350 | 0.290 | 0.342 |
| | 192 | **0.324** | **0.360** | 0.389 | 0.401 | 0.349 | 0.386 | 0.333 | 0.370 | 0.337 | 0.370 | 0.332 | 0.369 |
| | 336 | **0.357** | **0.379** | 0.393 | 0.414 | 0.377 | 0.404 | 0.369 | 0.392 | 0.369 | 0.385 | 0.366 | 0.392 |
| | 720 | **0.416** | **0.411** | 0.470 | 0.458 | 0.437 | 0.439 | **0.416** | 0.420 | 0.429 | 0.419 | 0.420 | 0.424 |
| ETTm2 | 96 | **0.160** | **0.251** | 0.201 | 0.280 | 0.176 | 0.270 | 0.166 | 0.256 | 0.163 | 0.252 | 0.167 | 0.260 |
| | 192 | **0.212** | **0.287** | 0.242 | 0.313 | 0.246 | 0.314 | 0.223 | 0.296 | 0.218 | 0.290 | 0.224 | 0.303 |
| | 336 | **0.261** | **0.322** | 0.310 | 0.356 | 0.304 | 0.352 | 0.274 | 0.329 | 0.274 | 0.330 | 0.281 | 0.342 |
| | 720 | **0.359** | **0.381** | 0.381 | 0.396 | 0.376 | 0.397 | 0.362 | 0.385 | 0.375 | 0.400 | 0.397 | 0.421 |
| ETTh1 | 96 | **0.362** | **0.390** | 0.389 | 0.412 | 0.396 | 0.421 | 0.379 | 0.401 | 0.372 | 0.399 | 0.375 | 0.399 |
| | 192 | **0.394** | **0.410** | 0.458 | 0.462 | 0.429 | 0.444 | 0.413 | 0.429 | 0.407 | 0.420 | 0.412 | 0.420 |
| | 336 | **0.419** | **0.426** | 0.523 | 0.501 | 0.464 | 0.469 | 0.435 | 0.436 | 0.426 | 0.433 | 0.439 | 0.443 |
| | 720 | **0.436** | **0.453** | 0.502 | 0.497 | 0.534 | 0.529 | 0.446 | 0.464 | 0.456 | 0.468 | 0.472 | 0.490 |
| ETTh2 | 96 | **0.267** | **0.332** | 0.338 | 0.397 | 0.307 | 0.362 | 0.274 | 0.337 | 0.273 | 0.339 | 0.289 | 0.353 |
| | 192 | **0.322** | **0.372** | 0.422 | 0.446 | 0.371 | 0.402 | 0.338 | 0.376 | 0.336 | 0.383 | 0.383 | 0.418 |
| | 336 | **0.351** | 0.401 | 0.431 | 0.460 | 0.425 | 0.439 | 0.363 | **0.397** | 0.360 | 0.405 | 0.448 | 0.465 |
| | 720 | **0.389** | 0.436 | 0.467 | 0.480 | 0.440 | 0.461 | 0.393 | **0.430** | 0.393 | **0.430** | 0.605 | 0.551 |
| Weather | 96 | **0.142** | **0.196** | 0.163 | 0.223 | 0.166 | 0.220 | 0.152 | 0.199 | 0.171 | 0.223 | 0.176 | 0.237 |
| | 192 | **0.186** | **0.239** | 0.218 | 0.266 | 0.213 | 0.255 | 0.197 | 0.243 | 0.217 | 0.266 | 0.192 | 0.246 |
| | 336 | **0.235** | **0.280** | 0.280 | 0.306 | 0.255 | 0.291 | 0.249 | 0.283 | 0.260 | 0.293 | 0.240 | 0.287 |
| | 720 | **0.310** | 0.336 | 0.349 | 0.356 | 0.325 | 0.340 | 0.320 | 0.335 | 0.326 | 0.348 | 0.316 | 0.352 |
| Traffic | 96 | 0.393 | 0.272 | 0.603 | 0.328 | **0.358** | 0.261 | 0.367 | **0.251** | 0.404 | 0.289 | 0.410 | 0.282 |
| | 192 | 0.407 | 0.279 | 0.610 | 0.329 | **0.379** | 0.269 | 0.385 | **0.259** | 0.407 | 0.289 | 0.423 | 0.287 |
| | 336 | 0.416 | 0.283 | 0.619 | 0.330 | **0.391** | 0.275 | 0.398 | **0.265** | 0.423 | 0.300 | 0.436 | 0.296 |
| | 720 | 0.453 | 0.305 | 0.632 | 0.352 | **0.426** | 0.293 | 0.434 | **0.287** | 0.462 | 0.310 | 0.466 | 0.315 |
| ECL | 96 | **0.130** | 0.225 | 0.181 | 0.281 | 0.132 | 0.229 | **0.130** | **0.222** | 0.140 | 0.239 | 0.140 | 0.237 |
| | 192 | **0.147** | 0.240 | 0.193 | 0.293 | 0.154 | 0.251 | 0.148 | **0.240** | 0.150 | 0.249 | 0.153 | 0.249 |
| | 336 | **0.163** | **0.257** | 0.205 | 0.312 | 0.165 | 0.263 | 0.167 | 0.261 | 0.170 | 0.266 | 0.169 | 0.267 |
| | 720 | **0.198** | **0.290** | 0.222 | 0.320 | 0.200 | 0.292 | 0.202 | 0.291 | 0.206 | 0.310 | 0.203 | 0.301 |
| Exchange | 96 | **0.079** | **0.199** | 0.160 | 0.296 | 0.124 | 0.257 | 0.087 | 0.209 | 0.086 | 0.212 | 0.086 | 0.208 |
| | 192 | **0.155** | **0.286** | 0.380 | 0.451 | 0.243 | 0.358 | 0.190 | 0.312 | 0.191 | 0.325 | 0.163 | 0.299 |
| | 336 | **0.279** | **0.385** | 0.659 | 0.615 | 0.385 | 0.461 | 0.377 | 0.446 | 0.309 | 0.414 | 0.311 | 0.424 |
| | 720 | **0.665** | **0.615** | 1.276 | 0.866 | 1.079 | 0.780 | 0.915 | 0.699 | 1.277 | 0.850 | 1.107 | 0.791 |

## 5. Ablation Studies

In this section, we aim to analyze the effectiveness of our designs for the TSF tasks. We instantiate our modules and test their performance on the non-stationary datasets.

**Table 3.** Univariate time series forecasting results. The input length is set as 336.

| Method | | WinNet | | TimesNet | | iTransformer | | PatchTST | | RLinear | | DLinear | |
|---|---|---|---|---|---|---|---|---|---|---|---|---|---|
| Metric | | MSE | MAE | MSE | MAE | MSE | MAE | MSE | MAE | MSE | MAE | MSE | MAE |
| ETTm1 | 96  | **0.025** | **0.121** | 0.028 | 0.126 | 0.031 | 0.134 | 0.026 | **0.121** | 0.029 | 0.127 | 0.028 | 0.123 |
|       | 192 | **0.038** | **0.148** | 0.048 | 0.167 | 0.051 | 0.173 | 0.039 | 0.150 | 0.043 | 0.154 | 0.045 | 0.156 |
|       | 336 | **0.051** | **0.171** | 0.060 | 0.188 | 0.066 | 0.197 | 0.053 | 0.173 | 0.064 | 0.187 | 0.061 | 0.182 |
|       | 720 | **0.062** | **0.189** | 0.076 | 0.213 | 0.075 | 0.201 | 0.074 | 0.207 | 0.081 | 0.212 | 0.080 | 0.210 |
| ETTm2 | 96  | **0.063** | 0.184 | 0.076 | 0.206 | 0.075 | 0.206 | 0.065 | 0.186 | 0.067 | 0.193 | **0.063** | **0.183** |
|       | 192 | **0.090** | **0.224** | 0.107 | 0.251 | 0.113 | 0.258 | 0.094 | 0.231 | 0.095 | 0.233 | 0.092 | 0.227 |
|       | 336 | **0.116** | **0.258** | 0.135 | 0.284 | 0.146 | 0.294 | 0.120 | 0.265 | 0.122 | 0.266 | 0.119 | 0.261 |
|       | 720 | **0.169** | **0.318** | 0.210 | 0.362 | 0.183 | 0.343 | 0.171 | 0.322 | 0.173 | 0.320 | 0.175 | 0.320 |
| ETTh1 | 96  | **0.052** | **0.176** | 0.062 | 0.195 | 0.060 | 0.188 | 0.055 | 0.179 | 0.059 | 0.183 | 0.056 | 0.180 |
|       | 192 | **0.068** | **0.203** | 0.080 | 0.225 | 0.072 | 0.207 | 0.071 | 0.205 | 0.078 | 0.212 | 0.071 | 0.204 |
|       | 336 | 0.080 | 0.225 | **0.075** | **0.215** | 0.080 | 0.222 | 0.081 | 0.225 | 0.100 | 0.248 | 0.098 | 0.244 |
|       | 720 | **0.079** | **0.225** | 0.079 | 0.225 | 0.079 | 0.225 | 0.087 | 0.232 | 0.181 | 0.351 | 0.189 | 0.359 |
| ETTh2 | 96  | **0.128** | **0.277** | 0.151 | 0.310 | 0.145 | 0.300 | 0.129 | 0.282 | 0.136 | 0.287 | 0.131 | 0.279 |
|       | 192 | **0.168** | **0.324** | 0.179 | 0.337 | 0.185 | 0.341 | 0.168 | 0.328 | 0.178 | 0.333 | 0.176 | 0.329 |
|       | 336 | 0.194 | 0.355 | 0.195 | 0.356 | 0.190 | **0.348** | **0.185** | 0.351 | 0.213 | 0.372 | 0.209 | 0.367 |
|       | 720 | 0.222 | 0.380 | 0.195 | 0.363 | **0.184** | **0.349** | 0.224 | 0.383 | 0.293 | 0.442 | 0.276 | 0.426 |

**Table 4.** Ablation study of our proposed modules in WinNet.

| Method | | Final | | TDD+DCB | | TDD | | Original | | TimesNet | |
|---|---|---|---|---|---|---|---|---|---|---|---|
| Metric | | MSE | MAE | MSE | MAE | MSE | MAE | MSE | MAE | MSE | MAE |
| Weather | 96  | **0.142** | **0.196** | 0.147 | 0.203 | 0.147 | 0.206 | 0.147 | 0.208 | 0.163 | 0.223 |
|         | 192 | **0.186** | **0.239** | 0.191 | 0.244 | 0.191 | 0.244 | 0.193 | 0.253 | 0.218 | 0.266 |
|         | 336 | **0.235** | **0.280** | 0.240 | 0.290 | 0.245 | 0.290 | 0.246 | 0.300 | 0.280 | 0.306 |
|         | 720 | **0.310** | **0.336** | 0.315 | 0.340 | 0.321 | 0.342 | 0.326 | 0.357 | 0.349 | 0.356 |
| ECL     | 96  | **0.130** | **0.225** | 0.135 | 0.231 | 0.139 | 0.237 | 0.145 | 0.249 | 0.181 | 0.281 |
|         | 192 | **0.147** | **0.240** | 0.150 | 0.244 | 0.153 | 0.249 | 0.163 | 0.266 | 0.193 | 0.293 |
|         | 336 | **0.163** | **0.257** | 0.167 | 0.261 | 0.169 | 0.264 | 0.180 | 0.283 | 0.205 | 0.312 |
|         | 720 | **0.198** | **0.290** | 0.211 | 0.300 | 0.208 | 0.297 | 0.217 | 0.315 | 0.222 | 0.320 |
| Traffic | 96  | **0.393** | **0.272** | 0.405 | 0.281 | 0.414 | 0.288 | 0.528 | 0.300 | 0.603 | 0.328 |
|         | 192 | **0.407** | **0.279** | 0.420 | 0.287 | 0.428 | 0.294 | 0.549 | 0.312 | 0.610 | 0.329 |
|         | 336 | **0.416** | **0.283** | 0.437 | 0.295 | 0.442 | 0.302 | 0.569 | 0.323 | 0.619 | 0.330 |
|         | 720 | **0.453** | **0.305** | 0.465 | 0.312 | 0.468 | 0.316 | 0.610 | 0.340 | 0.632 | 0.352 |

### 5.1  Model Architectures

To validate the proposed modules of WinNet, the ablation studies are conducted to determine the best model architecture, including Dual-Forecasting Heads, TDD and DCB blocks. TimesNet [6] is the SOTA benchmark for the CNN-based models. From **Table 4**, the original means series decomposition and a regular CNN network as the baseline and the final is our complete frame-work. In the original version, original series decomposition and a regular CNN network are applied to replace the TDD block, respectively. The original version fails to capture the long-term variation in complex datasets and suffers from inferior performance, however, it still outperforms the CNN-based method TimesNet [6]. From the traffic dataset, the TDD block can improve the MSE by 22.3%. Other modules can also contribute to expected performance improvements and finally outperform the chosen baselines.

### 5.2  Dual-Forecasting Heads

Table 5. Ablation study of Dual-Forecasting Heads.

| Method | | WinNet | | | | | | TimesNet | |
|---|---|---|---|---|---|---|---|---|---|
| | | Dual-Head | | Only Cross | | Only Within | | | |
| Metric | | MSE | MAE | MSE | MAE | MSE | MAE | MSE | MAE |
| Weather | 96 | **0.142** | **0.196** | 0.144 | **0.196** | 0.147 | 0.203 | 0.163 | 0.223 |
| | 192 | **0.186** | **0.239** | 0.188 | 0.242 | 0.191 | 0.244 | 0.218 | 0.266 |
| | 336 | **0.235** | **0.280** | 0.238 | 0.283 | 0.240 | 0.290 | 0.280 | 0.306 |
| | 720 | **0.310** | **0.336** | **0.310** | 0.339 | 0.315 | 0.340 | 0.349 | 0.356 |
| ECL | 96 | **0.130** | **0.225** | 0.137 | 0.235 | 0.135 | 0.231 | 0.181 | 0.281 |
| | 192 | **0.147** | **0.240** | 0.154 | 0.251 | 0.150 | 0.244 | 0.193 | 0.293 |
| | 336 | **0.163** | **0.257** | 0.171 | 0.268 | 0.167 | 0.261 | 0.205 | 0.312 |
| | 720 | **0.198** | **0.290** | 0.210 | 0.300 | 0.211 | 0.300 | 0.222 | 0.320 |
| Traffic | 96 | **0.393** | **0.272** | 0.413 | 0.301 | 0.405 | 0.281 | 0.603 | 0.328 |
| | 192 | **0.407** | **0.279** | 0.440 | 0.322 | 0.420 | 0.287 | 0.610 | 0.329 |
| | 336 | **0.416** | **0.283** | 0.436 | 0.311 | 0.437 | 0.295 | 0.619 | 0.330 |
| | 720 | **0.453** | **0.305** | 0.487 | 0.345 | 0.465 | 0.312 | 0.632 | 0.352 |

In WinNet, we design the two inputs, i.e., within- and cross-windows. We conduct an ablation study to validate the dual-forecasting mechanism on the model prediction ability. From **Table 5**, it can be seen that a single head can achieve more comparable results than TimesNet [6]. However, the single head is less effective on multi-channel large datasets like ECL and Traffic. The utilization of the Dual-Forecasting Heads yields a more effective effect in comparison to a single head. Therefore, more branched parallel design can provide the potential for the model to enhance prediction performance in multivariate time series forecasting tasks.

### 5.3  Sub-window Size

The size of the sub-window can also have implications for the experimental performance of the proposed model. We consider the model performance with different sub-window sizes to {18, 24, 32}, respectively. As demonstrated in **Table 6**, the proposed approach can achieve better performance for the sub-window size of 24 due to the lag correlation.

### 5.4   Input Length

In general, we consider that a larger look-back window can model more temporal dependencies and capture long- and short-term variations. To validate the performance of WinNet, the look-back windows are set as L = {24, 48, 96, 144, 192, 336, 512, 720}, and the prediction length T = 720. As shown in **Fig. 5**, WinNet can outperform the current SOTA models as the input length gets larger. From the figure, WinNet achieves slightly inferior performance than RLinear or TimesNet at 24 and 48 input lengths, due to the inability to capture long-term variations.

**Table 6.** prediction results with different sizes of the sub-window.

| Method | | WinNet | | | | | | TimesNet | |
|---|---|---|---|---|---|---|---|---|---|
| | | 18 | | 24 | | 32 | | | |
| Metric | | MSE | MAE | MSE | MAE | MSE | MAE | MSE | MAE |
| ETTh2 | 96 | 0.271 | 0.333 | **0.267** | **0.332** | 0.275 | 0.337 | 0.338 | 0.397 |
| | 192 | 0.326 | 0.374 | **0.322** | **0.372** | 0.337 | 0.377 | 0.422 | 0.446 |
| | 336 | 0.356 | **0.400** | **0.351** | 0.401 | 0.368 | 0.414 | 0.431 | 0.460 |
| | 720 | 0.395 | 0.437 | **0.389** | **0.436** | 0.406 | 0.448 | 0.467 | 0.480 |
| Weather | 96 | 0.151 | 0.207 | **0.142** | **0.196** | 0.146 | 0.202 | 0.163 | 0.223 |
| | 192 | 0.196 | 0.253 | **0.186** | **0.239** | 0.191 | 0.245 | 0.218 | 0.266 |
| | 336 | 0.245 | 0.288 | **0.235** | **0.280** | 0.239 | 0.289 | 0.280 | 0.306 |
| | 720 | 0.319 | 0.341 | **0.310** | **0.336** | 0.315 | 0.345 | 0.349 | 0.356 |
| ECL | 96 | 0.141 | 0.238 | **0.130** | **0.225** | 0.142 | 0.240 | 0.181 | 0.281 |
| | 192 | 0.155 | 0.251 | **0.147** | **0.240** | 0.157 | 0.253 | 0.193 | 0.293 |
| | 336 | 0.172 | 0.268 | **0.163** | **0.257** | 0.174 | 0.270 | 0.205 | 0.312 |
| | 720 | 0.211 | 0.300 | **0.198** | **0.290** | 0.212 | 0.301 | 0.222 | 0.320 |

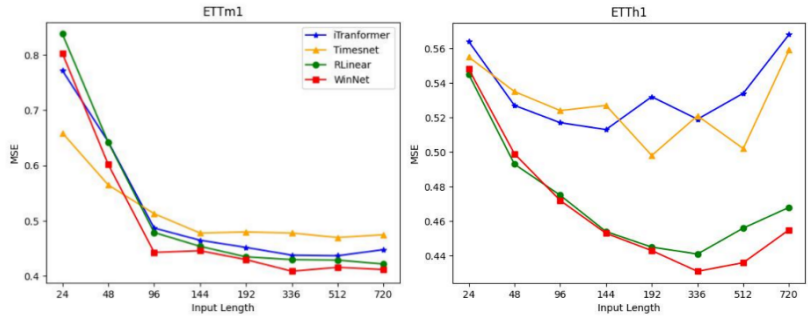

**Fig. 5.** The prediction results (MSE) with different look-back windows on ETTh1 and ETTm1.

### 5.5   Generalizability of TDD

To verify the generalizability of the proposed TDD block, we select MLP-based models (TiDE [13] and LightTS [9]) as the backbone models. As shown in **Table 7**, the TDD block can outperform the MLP-based models in almost all the datasets.

Table 7. Multivariate time series forecasting results of LightTS [9] and TiDE [13] with TDD.

| Method | | TiDE | | +TDD | | LightTS | | +TDD | |
|---|---|---|---|---|---|---|---|---|---|
| Metric | | MSE | MAE | MSE | MAE | MSE | MAE | MSE | MAE |
| ETTh1 | 96 | 0.388 | 0.396 | **0.387** | **0.395** | 0.424 | 0.432 | **0.391** | **0.404** |
| | 192 | 0.439 | 0.425 | **0.438** | **0.424** | 0.475 | 0.462 | **0.440** | **0.438** |
| | 336 | 0.485 | 0.450 | **0.482** | **0.448** | 0.518 | 0.488 | **0.484** | **0.460** |
| | 720 | 0.556 | 0.562 | **0.484** | **0.472** | 0.547 | 0.533 | **0.508** | **0.504** |
| ETTh2 | 96 | 0.400 | 0.440 | **0.291** | **0.340** | 0.397 | 0.437 | **0.337** | **0.392** |
| | 192 | 0.528 | 0.509 | **0.377** | **0.392** | 0.520 | 0.504 | **0.457** | **0.466** |
| | 336 | 0.643 | 0.571 | **0.417** | **0.427** | 0.626 | 0.559 | **0.582** | **0.535** |
| | 720 | 0.874 | 0.679 | **0.423** | **0.441** | 0.863 | 0.672 | **0.793** | **0.639** |
| Weather | 96 | 0.202 | 0.261 | **0.194** | **0.233** | 0.182 | 0.242 | **0.170** | **0.244** |
| | 192 | 0.242 | 0.290 | **0.238** | **0.269** | 0.227 | 0.287 | **0.222** | **0.289** |
| | 336 | **0.287** | 0.335 | 0.291 | **0.306** | 0.282 | 0.334 | **0.273** | **0.329** |
| | 720 | **0.351** | 0.386 | 0.364 | **0.352** | 0.352 | 0.386 | **0.322** | **0.367** |

### 5.6 Model Efficiency

In addition to the expected performance improvements, we also harvest a higher computational efficiency. **Table 8** shows the computational efficiency of our model in the univariate prediction tasks. See relevant computational efficiency by thop, torchsummary, and torch.cuda.memory_allocated functions. From the table, we can see that our model can achieve higher efficiency, in terms of computational complexity, number of parameters, memory consumption and inference time, even over the simple DLinear [5]. WinNet demonstrates superior lightweight performance, with reductions in parameters of 88.2% and 20.1% compared to iTransformer [11] and DLinear [5], respectively.

Table 8. Efficiency of our model on the Traffic dataset vs. other methods in univariate prediction task. We set the input and prediction length as 720. * means former.

| Method | WinNet | iTrans* | PatchTST | TimesNet | MICN | Cross* | DLinear |
|---|---|---|---|---|---|---|---|
| FLOPs | **851.3K** | 35.2M | 44.2M | 3240.7G | 5.32G | 726.5M | 1.04M |
| Params | **830.8K** | 7.05M | 8.69M | 450.9M | 18.75M | 11.09M | 1.04M |
| Memory | **11MiB** | 35MiB | 44MiB | 1762MiB | 85MiB | 56MiB | 12MiB |
| Training | 8ms | 11ms | 24ms | 491ms | 25ms | 55ms | **3ms** |
| Inference | **1.3ms** | 9.0ms | 11.2ms | 66.9ms | 12.7ms | 21.2ms | **1.3ms** |

## 6. Conclusion

In summary, we propose a CNN-based approach for time series forecasting task by introducing the TDD and DCB block. WinNet can eliminate the non-stationarity of the time series and capture the correlation between short- and long-term variations. We apply the TDD block to MLP-based methods and also achieve superior prediction performance. The correlation between trend and seasonal terms can provide the local variation. WinNet not only outperforms other baselines in terms of prediction accuracy but also harvests higher computational efficiency. This work demonstrates the potential for CNN-based methods in the TSF tasks.

**Acknowledgments.** This work was supported by National Natural Science Foundation of China (U2333209，62371323，U20A20161).